\documentclass[11pt,a4paper]{article}
\usepackage[hyperref]{naaclhlt2019}
\usepackage{times}
\usepackage{latexsym}

\usepackage{url}

\aclfinalcopy 
\def\aclpaperid{5} 

\usepackage[utf8]{inputenc} 
\usepackage[T1]{fontenc}    
\usepackage{booktabs}       
\usepackage{amsfonts}       
\usepackage{nicefrac}       
\usepackage{microtype}      
\usepackage{pgfplots,pgfplotstable}
\usepackage{subcaption}
\usepackage{multirow}

\title{Syntactic Interchangeability in Word Embedding Models}

\author{
Daniel Hershcovich \qquad Assaf Toledo \qquad Alon Halfon \qquad Noam Slonim \\
IBM Research\\
\texttt{daniel.hershcovich@gmail.com},\\
\texttt{assaf.toledo@ibm.com},\\
\texttt{\{alonhal,noams\}@il.ibm.com}
}

\begin{document}
    \maketitle
    
\setlength{\abovecaptionskip}{16pt}
\def\arraystretch{1.7}

    \begin{abstract}
    Nearest neighbors in word embedding models are commonly observed to be
    semantically similar, but the relations between them can vary greatly.
    We investigate the extent to which word embedding models
    preserve syntactic interchangeability, as reflected by distances between
    word vectors, and the effect of hyper-parameters---context window size in particular.
    We use part of speech (POS) as a proxy for syntactic interchangeability,
    as generally speaking, words with the same POS are syntactically valid in the same contexts.
    We also investigate the relationship between interchangeability
    and similarity as judged by commonly-used word similarity benchmarks,
    and correlate the result with the performance of word embedding models
    on these benchmarks.
    Our results will inform future research and applications in the selection
    of word embedding model, suggesting a principle for an appropriate selection
    of the context window size parameter depending on the use-case.
    \end{abstract}

    \section{Introduction}\label{sec:introduction}

    Word embedding algorithms \cite{mikolov2013efficient,pennington2014glove,levy2015improving}
    attempt to capture the semantic space of words
    in a metric space of real-valued vectors.
    While it is common knowledge that the hyper-parameters used to train these
    models affects the semantic properties of the distances arising from them
    \cite{bansal2014tailoring,lin2015unsupervised,goldberg2016primer,W17-0239},
    and indeed, it has been shown that
    they capture many different semantic relations \cite{yang2006verb,agirre2009study},
    little has been done to \textit{quantify} the
    effect of model hyper-parameters on output tendencies.
    Here we begin to answer this question,
    evaluating fastText \cite{bojanowski2016enriching}
    on benchmarks designed to measure how well a model
    captures the degree of similarity between words (\S\ref{sec:benchmarks}).
    
    In our experiments, we investigate how
    \textit{syntactic interchangeability} of words,
    represented by their part of speech (\S\ref{sec:interchangeability}),
    is expressed in word embedding models and evaluation benchmarks.
    
    Based on the distributional hypothesis \cite{harris1954distributional},
    word embeddings are learned from text
    by first extracting co-occurrences---finding, for each word token, all
    words within a context window around it,
    whose size (or maximal size) is a hyper-parameter of the training algorithm.
    Word vectors are then learned by predicting these co-occurrences or
    factorizing a co-occurrence matrix.

    We discover a clear relationship between the context window size hyper-parameter
    and the performance of a word embedding model in estimating the similarity between words.
    To try to explain this relationship, we quantify how syntactic interchangeability
    is reflected in each benchmark,
    and its relation to the context window size.
    Our experiments reveal that context window size is negatively correlated
    with the number of same-POS words among the nearest neighbors of words,
    but that this fact is not enough to explain the complex interaction between
    context window size and performance on word similarity
    benchmarks.\footnote{Our code and data are available at \url{https://github.com/danielhers/interchangeability}.}
    
    \begin{figure*}[th]
        \begin{subfigure}[b]{\columnwidth}
        \includegraphics[width=\columnwidth]{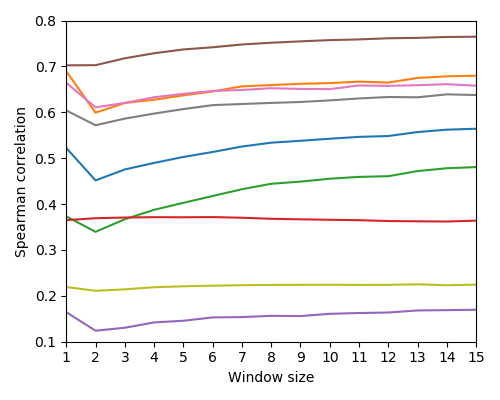}
        \caption{CBOW}
        \end{subfigure}
        \hfill
        \begin{subfigure}[b]{\columnwidth}
        \includegraphics[width=\columnwidth]{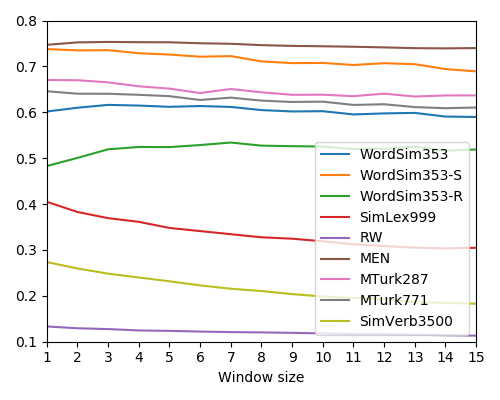}
        \caption{SGNS}
        \end{subfigure}
        \caption{Performance of the CBOW~(a) and SGNS~(b) algorithms on each benchmark,
        for each window size,
        measured by Spearman correlation between the benchmark score
        and the word embedding cosine similarity.\label{fig:benchmark_correlation}}
    \end{figure*}
%
    
    \section{Word Similarity and Relatedness}\label{sec:benchmarks}
    
    Many benchmarks have been proposed for the evaluation of unsupervised word
    representations.
    In general, they can be divided into intrinsic and extrinsic evaluation methods
    \cite{schnabel2015evaluation,chiu2016intrinsic,jastrzebski2017evaluate,alshargi2018concept2vec,bakarov2018survey}.
    While most datasets report the semantic similarity between words,
    many datasets actually capture semantic relatedness
    \cite{hill2015simlex,avraham2016improving},
    or more complex relations such as analogy or the ability to categorize
    words based on the distributed representation encoded in word embeddings.
    We focus on similarity and relatedness, and evaluate word embedding models
    on several common benchmarks.
    
    \subsection{Data}\label{sec:data}
    We learn word embeddings from English Wikipedia,
    using a dump from May 1, 2017.\footnote{\url{https://dumps.wikimedia.org/enwiki}}
    The data is preprocessed using a publicly available preprocessing
    script,\footnote{\url{http://mattmahoney.net/dc/textdata.html}}
    extracting text, removing nonalphanumeric characters,
    converting digits to text, and lowercasing the text.
    
    \paragraph{Benchmarks.}
    
    We use the following benchmarks:
     WordSim-353~\cite{finkelstein2001placing} and its partition into
     WordSim-353-Sim~\cite{agirre2009study} and
     WordSim-353-Rel~\cite{zesch2008using},
     SimLex999~\cite{hill2015simlex},
     Rare Words~\cite[RW; ][]{luong2013better},
     MEN~\cite{bruni2012distributional},
     MTurk-287~\cite{radinsky2011word},
     MTurk-771~\cite{halawi2012large}, and
     SimVerb-3500~\cite{Gerz2016emnlp}.
    See Table~\ref{tab:benchmark_enrichment} for the size of each benchmark.
    
    \subsection{Hyper-parameters}\label{sec:hyperparams}
    We use fastText \cite{bojanowski2016enriching} to learn
    300-dimensional word embedding models,
    using both the CBOW (continuous bag-of-words) and
    SGNS (skip-gram with negative sampling) algorithms \cite{mikolov2013efficient}.
    The context window size varies from 1 up to 15.
    We include only all words occurring 500 times or more
    (including function words), to avoid
    very rare words or uncommon spelling errors from skewing the results.
    All other hyper-parameters are set to their default values.
    
    \begin{table*}[th]
    \centering
    \begin{tabular}{@{}l|c||cc||cc|cc|c@{}}
    && \multicolumn{2}{c||}{\bf $\Delta \mathtt{win}=2\to 15 (\%)$}
    & \multicolumn{2}{c|}{\bf $\#$ Related} & \multicolumn{2}{c|}{\bf $\#$ Unrelated} \\
    \bf Benchmark & \bf Size& \bf CBOW & \bf SGNS
    & \bf All & \bf Same-POS & \bf All & \bf Same-POS & \bf p-value \\
    \hline
    WordSim353 & 353 & 24 & -3 & 122 & 107 & 53 & 40 & 0.038 \\
    WordSim353-S & 203 & 13 & -6 & 60 & 53 & 53 & 40 & 0.061 \\
    WordSim353-R & 252 & 42 & 4 & 104 & 89 & 39 & 31 & 0.26 \\
    SimLex999 & 999 & -1 & -20 & 234 & 199 & 334 & 295 & 0.897 \\
    RW & 2034 & 37 & -12 & 944 & 555 & 262 & 144 & 0.149 \\
    MEN & 3000 & 9 & -2 & 791 & 564 & 781 & 439 & $3\cdot10^{-10}$ \\
    MTurk287 & 287 & 8 & -5 & 49 & 39 & 119 & 68 & 0.004 \\
    MTurk771 & 771 & 12 & -5 & 204 & 153 & 200 & 146 & 0.365 \\
    SimVerb3500 & 3500 & 6 & -30 & 633 & 265 & 1217 & 566 & 0.974
    \end{tabular}
    \caption{Analysis of interchangeability (by same-POS) in
    word similarity and relatedness benchmarks.
    $\Delta \mathtt{win}=2\to 15 (\%)$ is the relative change, in percents,
    of the model's performance (by Spearman correlation) when going from window size 2
    to window size 15, for the CBOW
    and SGNS algorithms (\S\ref{sec:eval_exp}).
    \textit{Related} and \textit{Unrelated} are the top and bottom 30\% of the pairs,
    by benchmark score, respectively.
    \textit{P-value} is calculated using the hypergeometric
    test, comparing the enrichment of interchangeable pairs within related pairs,
    with a background of all related and unrelated pairs (\S\ref{sec:benchmark_exp}).\label{tab:benchmark_enrichment}}
    \end{table*}
    
    \subsection{Evaluation on Benchmarks}\label{sec:eval_exp}
    
    To investigate the effect of window size on a model's performance on the benchmarks,
    we evaluate each model on each benchmark, using cosine similarity
    as the model's prediction for each pair.
    The performance is measured by Spearman correlation between the benchmark score
    and the word embedding cosine similarity \cite{levy2015improving}.
    
    \paragraph{Results.}
    
    Figure~\ref{fig:benchmark_correlation} displays the performance of the CBOW and SGNS
    algorithms on each benchmark, with window sizes 1 to 15.
    Apart from a small dip between windows 1 and 2 for
    CBOW,
    the performance is either nearly constant,
    or changes nearly monotonically with window size in each setting.
    
    The relative improvement (or deterioration),
    in percents, with the increase of window size from 2 to 15,
    are shown in Table~\ref{tab:benchmark_enrichment} ($\Delta \mathtt{win}=2\to 15 (\%)$).
    Interestingly, CBOW exhibits a positive correlation of window size with model's performance
    for all benchmarks but SimLex999,
    while performance for SGNS barely changes with window size,
    except for SimLex999 and SimVerb3500, where we see a strong \textit{negative} correlation.
    
    \paragraph{Discussion.}
    
    In SimLex999 and in SimVerb3500, the words in each pair have the same part of speech by design
    (in particular, SimVerb3500 only contains verbs).
    Hypothesizing that the effect of window size is related to the model's
    implicitly learned concept of part of speech, we investigate this idea in the next section.

    \section{Syntactic Interchangeability}\label{sec:interchangeability}
    
    A word's part of speech (also known as syntactic category)
    is determined by syntactic distribution, and
    conveys information about how a word functions in the sentence \cite{carnie2002syntax}.
    We can generally substitute each word in a sentence
    with various words that are of the same part of speech,
    but not words that are of different parts of speech.
    While the same syntactic function can sometimes be fulfilled by words of
    various parts of speech or possibly longer phrases (such as adverbs and
    prepositional phrases, or multi-word expressions),
    part of speech is nonetheless a very good proxy for syntactic distribution
    \cite{W04-2404}.
    
    Related to our work,
    \newcite{K17-1013} introduced a framework for automatic selection of
    specific context configurations for word embedding models
    per part of speech, improving
    performance on the SimLex999 benchmark.
    We take a different approach, investigating existing word embedding
    models and the way in which part of speech is reflected in them.
    
    We define two words to be (syntactically) \textit{interchangeable}
    if they share the same part of speech.
    We quantify interchangeability
    as a property of a word embedding model,
    as the proportion of words with the same part of speech
    within the list of nearest neighbors
    (that is, the most similar words according to the model)
    for each word in a pre-determined vocabulary.
    The higher the interchangeability ratio is,
    the more importance we assume the model implicitly places on interchangeability
    for the calculation of word similarity.
    
    \subsection{Interchangeability Analysis in Word Similarity Benchmarks}\label{sec:benchmark_exp}
    
    While all benchmarks we experiment with assign a score along a scale to each pair
    (calculated from human scoring), for our experiment we would like to use
    a binary annotation of whether a pair is related or not.
    For this purpose, we divide the whole range of scores,
    for each benchmark, to three parts:
    the lowest 30\% of the range between the lowest and highest scores
    is considered ``unrelated'', the top 30\% as ``related'',
    and the middle 40\% are ignored.
    
    \paragraph{Interchangeability enrichment.}
    
    Given the binary classification obtained from the human-annotated scores
    for each benchmark, we can find the enrichment of interchangeable pairs among
    related pairs.
    We use spaCy 2.0.11\footnote{\url{https://spacy.io}} (with the \texttt{en\_core\_web\_sm} model)
    to annotate the POS for each word in each benchmark
    pair (tagging them in isolation to select the most probable POS),
    and look at the set of same-POS pairs in the benchmark.
    For each of the benchmarks, we calculate a p-value using the hypergeometric
    test, comparing the enrichment of same-POS pairs within related pairs,
    with a background distribution of all related and unrelated pairs (ignoring ones in
    the middle 40\% range of scores).
    
    \paragraph{Results.}
    
    Table~\ref{tab:benchmark_enrichment} shows the enrichment of interchangeable pairs
    among related and unrelated pairs for each benchmark.
    For WordSim353, MEN and MTurk287, the set of related pairs
    contains significantly more interchangeable pairs than the background
    set ($p<0.05$),\footnote{The fact that not all pairs in SimLex999 and SimVerb3500
    are judged as interchangeable
    in our experiment is due to ambiguity: for some words, spaCy selected a POS
    which is not the one intended when constructing the benchmark.}
    suggesting that these benchmarks are particularly sensitive to POS.

    \subsection{Nearest Neighbor Analysis}\label{sec:interchangeability_exp}
    
    To try and relate the results from \S\ref{sec:eval_exp} and \S\ref{sec:benchmark_exp},
    we measure the relation between window size and interchangeability
    by analyzing nearest neighbors in word embedding models.
    In our experiment, the \textit{nearest neighbors} of a word are the words
    with the highest cosine similarity between their vectors.
        
    \paragraph{Collecting pivots.}
    
    We create a word list for each of the three most
    common parts of speech:
    nouns, adjectives and verbs.
    For each POS, we list all lemmas of all synsets of that POS from
    WordNet \cite{miller1998wordnet}.
    To ``purify'' the lists and avoid noise from homonyms,
    we remove from each list any lemma that also belongs to a synset from
    another POS.
    As a further cleaning step, we use spaCy to tag each word,
    and only keep words for which the spaCy POS agrees with the WordNet POS.
    Without context, spaCy will likely choose the most
    common POS based on its training corpus, which is different from WordNet,
    increasing the robustness.
    
    This process results in 6407 \textit{uniquely-noun}, 2784 \textit{uniquely-adjective}
    and 1460 \textit{uniquely-verb} words, which we refer to as our \textit{pivot lists}.
    
    \paragraph{Calculating nearest neighbor POS.}
    
    We find the 100 nearest neighbors for each word in our pivot lists,
    according to each fastText model with windows 1 through 15.
    We filter these neighbors to
    keep only words in the spaCy vocabulary, and inspect the remaining top 10.
    Again using spaCy, we tag the POS of each neighbor in the result.
    We subsequently calculate a histogram, for each POS $x$, of its
    \textit{neighbor-POS} $y$, that is, the POS assigned to the neighbors of
    words with POS $x$.
    
    \begin{table}[t]
    \centering
    \setlength\tabcolsep{2.5pt}
    \begin{tabular}{l|ccc|ccc|ccc}
    {\bf Algo-} & \multicolumn{3}{c|}{\bf NOUN} & \multicolumn{3}{c|}{\bf ADJ} & \multicolumn{3}{c}{\bf VERB} \\
    {\bf rithm} & 1 & 15 & r & 1 & 15 & r & 1 & 15 & r \\
    \hline
    CBOW & 79 & 70 & -0.96 & 72 & 48 & -0.93 & 55 & 41 & -0.91 \\
    SGNS & 78 & 66 & -0.95 & 66 & 39 & -0.94 & 51 & 41 & -0.92 
    \end{tabular}
    \caption{Percentage of interchangeable neighbors per pivot POS for the smallest (1) and largest (15)
        windows in our experiment, for the CBOW and SGNS algorithms.
        The number of interchangeable neighbors has a strong negative Pearson correlation (r) with window size
        for windows 1 to 15 ($p<0.01$, two-tailed t-test).\label{tab:nn_pos_hist}}
    \end{table}
    
    \paragraph{Results.}
    
    Table~\ref{tab:nn_pos_hist} shows the results of this experiment.
    For nouns, adjectives and verbs, we consistently see a decrease in
    the  number of same-POS neighbors when we increase the window size,
    relative to the total number of nearest
    neighbors.

\begin{figure*}[t]
        \begin{subfigure}[b]{.475\textwidth}
        \includegraphics[width=\textwidth]{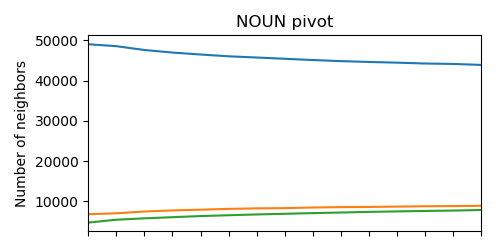}
        \includegraphics[width=\textwidth]{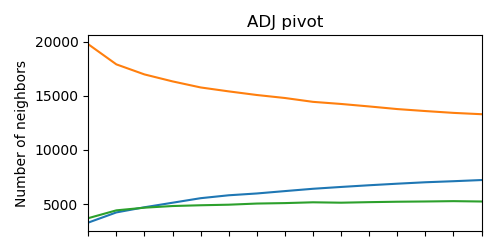}
        \includegraphics[width=\textwidth]{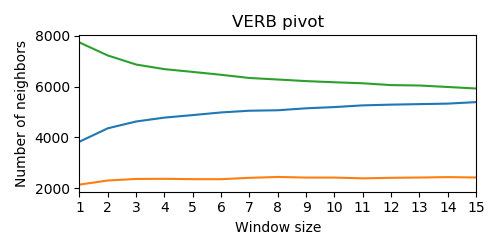}
        \caption{CBOW}
        \end{subfigure}
        \hfill
        \begin{subfigure}[b]{.475\textwidth}
        \includegraphics[width=\textwidth]{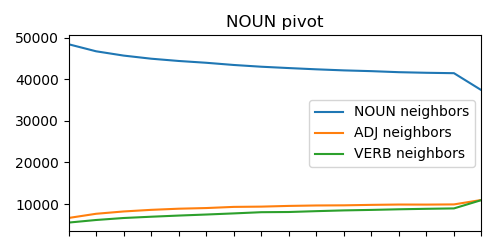}
        \includegraphics[width=\textwidth]{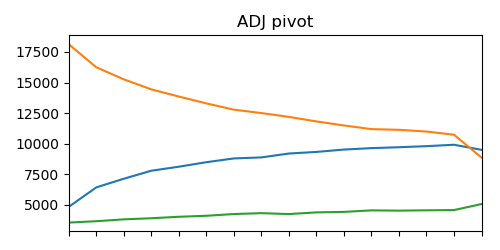}
        \includegraphics[width=\textwidth]{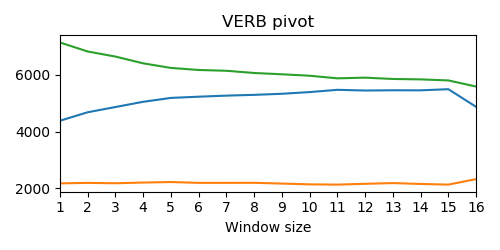}
        \caption{SGNS}
        \end{subfigure}
        \caption{Number of neighbor per POS for each pivot POS and for each window size,
        for the CBOW~(a) and SGNS~(b) algorithms.
        The number of same-POS neighbors is consistently decreasing with window size.\label{fig:nn_pos_hist}}
    \end{figure*}

    Figure~\ref{fig:nn_pos_hist} shows the the absolute number of neighbors per algorithm,
    pivot POS and neighbor POS, for all window sizes we experimented with.
    The number of nearest neighbors of the same POS is consistently decreasing with window size,
    while the number of nearest neighbors of other POS are increasing or unaffected.
    
    \paragraph{Discussion.}
    
    The results clearly suggest that for \textit{both} CBOW and SGNS,
    models with a larger window size are less likely to consider words
    of the same POS as strongly related.
    That is, syntactic interchangeability is negatively correlated with window size.
    This is in sharp contrast to our results from \S\ref{sec:eval_exp},
    where performance for CBOW on almost all benchmarks
    (among them  WordSim353, MEN and MTurk287, for which we showed that
    syntactic interchangeability plays a role) consistently
    \textit{improved} with window size.
    We also find the conclusion to contradict the impression regarding SGNS,
    where SimLex999 and SimVerb3500 showed worse performance for larger windows:
    if POS should not play a role in these benchmarks,
    then models with a bias toward syntactic interchangeability (i.e., models with lower windows)
    should perform \textit{worse} on these benchmarks.

\section{Conclusion}\label{sec:conclusion}

    We investigated the effect of the context window size hyper-parameter
    on the performance on word similarity benchmarks.
    We showed that
    (1) increasing the window size results in a lower probability of interchangeable
    nearest neighbors for both CBOW and SGNS algorithms;
    (2) in some widely used benchmarks,
    syntactic interchangeability increases the probability of similarity or relatedness;
    (3) increasing the window size typically improves performance
    in predicting similarity or relatedness for CBOW,
    but has little impact on SGNS.
    
    SimLex999 and SimVerb3500 proved to be exceptions to both (2) and (3),
    since all pairs in them are interchangeable by construction,
    but on them, increasing the window size has no effect for CBOW
    and negative impact for SGNS.
    
    This contradiction is presented as a challenge to the community,
    and could perhaps be explained by other factors affected by window size.
    
    Our investigation focused on a specific relation between words,
    namely whether they share a part of speech.
    Many other relations are of interest to the NLP
    community, such as syntactic dependency relations,
    and semantic relations like hypernymy and synonymy.
    Furthermore, a similar analysis could be applied to other word
    embedding hyper-parameters, such as the vector dimension.
    While we used a constant vector dimension of 300 in our experiments,
    it is an open question whether models with different vector dimensions
    differ with respect to their tendency to capture different word relations.
    Future work will extend our analysis to other relations and hyper-parameters.

\section*{Acknowledgments}

We thank the anonymous reviewers for their helpful comments.

    \bibliographystyle{acl_natbib}
    \bibliography{references}
    
\end{document}